\documentclass[12pt]{article}
\usepackage{graphicx} % Required for inserting images
\usepackage{xcolor}
\usepackage{amsmath}
\usepackage{algorithm}
\usepackage{amsmath}
\usepackage{amssymb}
\usepackage{indentfirst}
\usepackage{algpseudocode}
\usepackage{booktabs}
\usepackage{float}
\newtheorem{theorem}{Theorem}
\usepackage{caption}
\providecommand{\keywords}[1]
{
  \small	
  \textbf{Keywords:} #1
}

\title{Machine Learning Techniques for Multifactor Analysis of National Carbon Dioxide Emissions}

\title{Machine Learning Techniques for Multifactor Analysis of National Carbon Dioxide Emissions\thanks{Research supported by ADIALab}}

\author{Wenjia Xie\thanks{Department of Mathematical Sciences, Tsinghua University, Beijing, China, 100084. E-mail: jiewj20@mails.tsinghua.edu.cn},
Jinhui Li\thanks{Department of Mathematics, University of Toronto, Toronto, ON, Canada, M5T 3J1. E-mail: jinhuidavis.li@mail.utoronto.ca},
Kai Zong\thanks{China Academy of Safety Science and Technology, Beijing, China, 100012. E-mail: zongkai@hotmail.com
}, 
Luis Seco\thanks{Department of Mathematics, University of Toronto, Toronto, ON, Canada, M5T 3J1. E-mail: luis.seco@utoronto.ca}
}

\date{March 2025}

\begin{document}

\maketitle
\begin{abstract}

This paper presents a comprehensive study leveraging Support Vector Machine (SVM) regression and Principal Component Regression (PCR) to analyze carbon dioxide emissions in a global dataset of 62 countries and their dependence on idiosyncratic, country-specific parameters. 
The objective is to understand the factors contributing to carbon dioxide emissions and identify the most predictive elements. The analysis provides country-specific emission estimates, highlighting diverse national trajectories and pinpointing areas for targeted interventions in climate change mitigation, sustainable development, and the growing carbon credit markets and green finance sector. The study aims to support policymaking with accurate representations of carbon dioxide emissions, offering nuanced information for formulating effective strategies to address climate change while informing initiatives related to carbon trading and environmentally sustainable investments.
\end{abstract}


\section{Introduction}
Understanding the factors driving carbon dioxide emissions is essential for developing effective policies and fostering sustainable development. This study applies Support Vector Machine (SVM) regression and Principal Component Regression (PCR) to analyze a comprehensive dataset spanning 62 countries from 1992 to 2019. The analysis focuses on ten socioeconomic and environmental variables, including population, surface area, total fossil fuel consumption, electricity production, GDP, urban population, construction value, manufacturing, number of livestock, and agricultural gross production.

Extensive preprocessing ensured that the data were standardized and stationary, facilitating reliable analysis. The SVM regression model was fine-tuned through hyperparameter optimization to maximize its performance in capturing emissions-related patterns. Simultaneously, PCR was employed to address multicollinearity among variables, improving the stability and interpretability of the results.

The core of this research is an in-depth evaluation of the SVM and PCR models. By analyzing the alignment of the modeled emissions with actual data, the study identifies the influence of each variable using Permutation Importance. This approach provides valuable insights into the complex dynamics of carbon dioxide emissions and highlights key factors that drive variability across countries. These findings enable a more nuanced understanding of emissions at both global and national scales, offering actionable information for shaping policies aimed at mitigating climate change.

By combining advanced machine learning techniques and feature importance analysis, this study provides a robust framework for understanding carbon dioxide emissions. The integration of SVM and PCR not only enhances the reliability of the analysis but also underscores the interplay of socioeconomic and environmental factors. The ultimate aim is to support global sustainability efforts by equipping policymakers with data-driven insights into the determinants of emissions and identifying areas for targeted intervention and resource allocation.

\section{Literature Review}

The application of machine learning techniques has significantly advanced the analysis and estimation of carbon dioxide (CO\(_2\)) emissions, enabling researchers to manage complex datasets and identify underlying patterns. Various models have been employed to examine and understand CO\(_2\) emissions, highlighting the strengths of machine learning in addressing environmental challenges.

Kavoosi et al. utilized a genetic algorithm (GA) to estimate CO\(_2\) emissions, demonstrating the potential of evolutionary algorithms in environmental modeling \cite{kavoosi2012forecast}. Similarly, Sun applied an optimized grey forecasting model based on Harmony Search to examine emissions in China, showcasing the effectiveness of hybrid optimization techniques \cite{sun2013forecasting}. Abdel employed an Artificial Neural Network (ANN) model to analyze time series of CO\(_2\) emissions, highlighting the adaptability of neural networks in capturing nonlinear relationships \cite{baareh2013solving}.

Expanding the range of approaches, Kaboli et al. incorporated adaptive neuro-fuzzy inference systems (ANFIS), ANN, support vector regression (SVR), gene expression programming (GEP), particle swarm optimization (PSO), and backtracking search algorithms (BSA) to evaluate energy usage, a key driver of CO\(_2\) emissions \cite{kaboli2017long}. Lu et al. utilized a three-layer perceptron neural network to assess transportation-related CO\(_2\) emissions, emphasizing the role of deep learning in sector-specific analyses \cite{lu2017predicting}. Gholizadeh and Sabzi employed ANFIS and ANN techniques to analyze CO\(_2\) emissions \cite{gholizadeh2017prediction}, while Norhayati and Rashid used real data from medical waste incineration facilities for ANFIS-based CO\(_2\) emission assessments \cite{norhayati2018adaptive}.

Hybrid models combining ANFIS with optimization algorithms, such as PSO and genetic algorithms, have further enhanced the precision of energy and emissions analysis. These methodologies demonstrate how integrated approaches can refine the understanding of CO\(_2\) dynamics and their contributing factors.

Support vector machines (SVM) have also been extensively applied for time series analysis due to their robustness in handling nonlinear data. Mehdizadeh and Movagharnejad found that SVM surpassed semi-empirical models in estimating solute behavior in supercritical CO\(_2\) environments \cite{mehdizadeh2011comparison}. De Paz demonstrated that SVM's structural risk minimization principle effectively addresses CO\(_2\) exchange rates \cite{saleh2016carbon}. Similarly, Wang et al. identified SVM as an effective tool for analyzing time series in machine learning \cite{wang2008online}. Saleh et al. achieved high-accuracy estimations of CO\(_2\) emissions using SVM, providing valuable insights for emission analysis \cite{saleh2016carbon}.

Machine learning models have been increasingly employed to analyze CO\(_2\) emissions across diverse datasets. A study encompassing data from 1960 to the present demonstrated the value of machine learning in providing actionable insights for shaping environmental strategies and policies aimed at mitigating climate change \cite{zhang2024machine}.

In conclusion, the integration of machine learning techniques, including evolutionary algorithms, neural networks, and hybrid models, has significantly enhanced the ability to analyze CO\(_2\) emissions. These advancements provide researchers and policymakers with critical tools to develop effective strategies for managing carbon emissions and addressing their impact on climate change.
\section{Features}
Features are central to understanding complex phenomena such as carbon volume. This section examines the significance of ten key socioeconomic and environmental features—Total Fossil Fuel Consumption, GDP, Population, Urban Population, Electricity Production, Surface Area, Construction Value, Manufacturing, Number of Livestock, and Agriculture Gross Production—in relation to carbon emissions. Each feature is detailed in terms of its significance, quantification method, and typical value range. All data were sourced from the World Bank \cite{worldbankdata} and NationMaster database \cite{nationmasterdata}.

\begin{enumerate}
    \item \textbf{Total Fossil Fuel Consumption (GWh)}
    \begin{itemize}
        \item \textbf{Importance}: Critical for assessing a nation's carbon dioxide emissions footprint.
        \item \textbf{Quantification}: Combined fossil fuel energy consumption across all sectors annually.
        \item \textbf{Range}: From hundreds to hundreds of thousands of GWh.
        \item \textbf{Unit}: Gigawatt-hours (GWh).
    \end{itemize}
    
    \item \textbf{GDP (US \$)}
    \begin{itemize}
        \item \textbf{Importance}: Indicates economic activity level, correlating with carbon dioxide emissions.
        \item \textbf{Quantification}: Annual GDP in purchasing power parity or nominal values.
        \item \textbf{Range}: From billions to trillions.
        \item \textbf{Unit}: US Dollars (USD).
    \end{itemize}
    
    \item \textbf{Population}
    \begin{itemize}
        \item \textbf{Importance}: Directly influences carbon dioxide emissions through increased demand for energy and services.
        \item \textbf{Quantification}: Annual population figures.
        \item \textbf{Range}: From thousands to billions.
        \item \textbf{Unit}: Individuals.
    \end{itemize}
    
    \item \textbf{Urban Population}
    \begin{itemize}
        \item \textbf{Importance}: Urbanization increases energy consumption and carbon dioxide emissions.
        \item \textbf{Quantification}: Number of individuals living in urban areas.
        \item \textbf{Range}: From thousands to hundreds of millions.
        \item \textbf{Unit}: Individuals.
    \end{itemize}
    
    \item \textbf{Electricity Production (GWh)}
    \begin{itemize}
        \item \textbf{Importance}: Source and scale of electricity production impact carbon dioxide volumes.
        \item \textbf{Quantification}: Total annual electricity production.
        \item \textbf{Range}: From thousands to billions of GWh.
        \item \textbf{Unit}: Gigawatt-hours (GWh).
    \end{itemize}
    
    \item \textbf{Surface Area (Square KM)}
    \begin{itemize}
        \item \textbf{Importance}: Land use and size of a country influence carbon dioxide emissions.
        \item \textbf{Quantification}: Total land area.
        \item \textbf{Range}: From small areas to vast expanses.
        \item \textbf{Unit}: Square kilometers (km\(^2\)).
    \end{itemize}
    
    \item \textbf{Construction Value (US \$)}
    \begin{itemize}
        \item \textbf{Importance}: Reflects construction activity levels, tied to urban development and emissions.
        \item \textbf{Quantification}: Annual financial value of construction.
        \item \textbf{Range}: Varied, dependent on development stage.
        \item \textbf{Unit}: US Dollars (USD).
    \end{itemize}
    
    \item \textbf{Manufacturing (US \$)}
    \begin{itemize}
        \item \textbf{Importance}: Measures industrial production, a significant factor in energy use and emissions.
        \item \textbf{Quantification}: Monetary value of manufactured goods annually.
        \item \textbf{Range}: Reflects industrial capacity.
        \item \textbf{Unit}: US Dollars (USD).
    \end{itemize}
    
    \item \textbf{Number of Livestock (Heads)}
    \begin{itemize}
        \item \textbf{Importance}: Agricultural activity level indicator, contributing to methane and carbon dioxide emissions.
        \item \textbf{Quantification}: Total count of livestock.
        \item \textbf{Range}: Varies widely.
        \item \textbf{Unit}: Heads.
    \end{itemize}
    
    \item \textbf{Agriculture Gross Production (million US \$)}
    \begin{itemize}
        \item \textbf{Importance}: Economic output of agriculture, influencing land use and emissions.
        \item \textbf{Quantification}: Economic value of agricultural production.
        \item \textbf{Range}: Based on productivity and market value.
        \item \textbf{Unit}: Million US Dollars (USD).
    \end{itemize}
\end{enumerate}

Before implementing any procedures on our dataset, it is crucial to pre-process the raw feature data. The first step in this process is standardization, in which we rescale each feature so that it has a mean of 0 and a standard deviation of 1. This is performed irrespective of the country or year to which the data pertain. Following standardization, our initial plan was to apply differencing to achieve stationarity in the data for more reliable forecasting. However, upon conducting the Augmented Dickey-Fuller (ADF) test, we ascertained that our data are already stationary, rendering the differencing step unnecessary.

In detail, data stationarity implies a constant mean and variance over time, alongside a consistent covariance between different time intervals within the time series. This property of stationarity ensures the independence of data, a pivotal assumption for many statistical models. The Augmented Dickey-Fuller (ADF) test is a well-established statistical test for determining the stationarity of a given time series. It extends the Dickey-Fuller test by incorporating lagged differences to account for autocorrelation in the data.

The mathematical representation of the ADF test is as follows:
\begin{align*}
    \Delta y_t = \alpha + \beta t + \gamma y_{t-1} + \delta_1 \Delta y_{t-1} + \delta_2 \Delta y_{t-2} + \ldots + \delta_p \Delta y_{t-p} + \varepsilon_t,
\end{align*}
where \( y_t \) represents the time series data, \( \alpha \) is a constant, \( \beta t \) is the coefficient for a time trend, \( \gamma \) is the coefficient for \( y_{t-1} \), the lagged value of the time series, \( \delta_i \) are the coefficients for the lagged differences, \( p \) is the number of lags included in the model, and \( \varepsilon_t \) is the error term.

The coefficient \( \gamma \) is the focus of the test because it is associated with the lagged value of the time series \( y_{t-1} \). The null hypothesis \( H_0 \) posits that if \( \gamma = 0 \), it indicates that the time series has a unit root and is nonstationary. The alternative hypothesis posits that if \( \gamma < 0 \), it suggests that the time series does not have a unit root and is stationary.

\section{Methodology}
In this section, we outline the methodology employed to predict carbon volume and investigate the importance of factors that contribute to the prediction. We utilize a support vector machine (SVM) as the primary machine learning technique and apply Permutation Importance as a feature selection method to identify the most important factors. To address multicollinearity in our data, we incorporate Principal Component Analysis (PCA) and Principal Component Regression (PCR) to reduce dimensionality and ensure stable estimates of regression coefficients.

\subsection{Support Vector Regression (SVR)}

Support Vector Machine (SVM) is a powerful supervised learning algorithm that aims to find the optimal hyperplane in a high-dimensional feature space. For predicting carbon dioxide volume based on multiple factors, we employ support vector regression (SVR), which handles both linear and non-linear relationships. The primary objective of SVR is to find a function $f(x)$ that approximates the target variable $y$ as closely as possible, defined by $f(x) = \langle w, x \rangle + b$.

SVR uses a $\epsilon$ insensitive loss function, $L(y, f(x)) = \max(0, |y - f(x)| - \epsilon)$, to minimize the impact of errors within a certain margin. The optimization problem involves minimizing the following.

\begin{equation}
\min_{w, b, \xi, \xi^*} \frac{1}{2} ||w||^2 + C \sum_{i=1}^{n} (\xi_i + \xi_i^*),
\end{equation}
subject to the constraints:
\begin{equation}
y_i - \langle w, x_i \rangle - b \leq \epsilon + \xi_i,
\end{equation}
\begin{equation}
\langle w, x_i \rangle + b - y_i \leq \epsilon + \xi_i^*,
\end{equation}
\begin{equation}
\xi_i, \xi_i^* \geq 0.
\end{equation}

SVR can handle both linear and non-linear relationships, making it well suited for predicting carbon volume based on multiple factors. By maximizing the margin between the hyperplane and the support vectors, SVR seeks to find the best fit for the data, ensuring accurate predictions.

The primary objective of SVR is to find a function \( f(x) \) that approximates the target variable \( y \) as closely as possible. The function \( f(x) \) is defined as 
\begin{equation}
f(x) = \langle w, x \rangle + b,
\end{equation}
where \( \langle w, x \rangle \) is the dot product between the weight vector \( w \) and the feature vector \( x \), and \( b \) is the bias term.

SVR uses an \( \epsilon \)-insensitive loss function, which means the errors within a certain margin are ignored. The loss function \( L \) is defined as:
\begin{equation}
L(y, f(x)) = \max(0, |y - f(x)| - \epsilon).
\end{equation}

The optimization problem in SVR is to minimize the following objective function:
\begin{equation}
\min_{w, b, \xi, \xi^*} \frac{1}{2} ||w||^2 + C \sum_{i=1}^{n} (\xi_i + \xi_i^*),
\end{equation}
subject to the constraints:
\begin{equation}
y_i - \langle w, x_i \rangle - b \leq \epsilon + \xi_i,
\end{equation}
\begin{equation}
\langle w, x_i \rangle + b - y_i \leq \epsilon + \xi_i^*,
\end{equation}
\begin{equation}
\xi_i, \xi_i^* \geq 0.
\end{equation}

SVR can also be extended to solve nonlinear problems by applying the "kernel trick," which involves mapping the input features into a higher-dimensional space. The kernel function \( K(x, x') \) replaces the dot product \( \langle x, x' \rangle \) in the optimization problem.

Commonly used kernel functions include:
\begin{itemize}
    \item Linear: \( K(x, x') = \langle x, x' \rangle \),
    \item Polynomial: \( K(x, x') = (1 + \langle x, x' \rangle)^d \),
    \item Radial Basis Function (RBF): \( K(x, x') = \exp(-\gamma ||x - x'||^2) \).
\end{itemize}

\subsection{Permutation Importance}
To determine the relative importance of factors, we employ Permutation Importance as a feature selection technique. Permutation Importance measures the impact of permuting the values of each feature on the model's performance. By ranking the features through this process, we gain insights into the factors that have the most significant impact on the volume of carbon dioxide. This combined approach of SVR and Permutation Importance allows us to make accurate predictions while identifying the key drivers behind carbon emissions.

Given a predictive model \(M\) trained on tabular data \(D\) consisting of \(n\) features, and the performance score \(S\) of model \(M\) evaluated on dataset \(D\).

\begin{theorem}
For each feature \(j\) in the dataset \(D\), the importance value \(I_j\) is determined by the difference in the performance of the model when the feature \(j\) is used normally versus when the values of the feature \(j\) are randomly permuted. This importance value is calculated as follows:
\begin{equation}
I_j = S - \frac{1}{L} \sum_{l=1}^{L} S_j^l,
\end{equation}
where \(S_j^l\) is the performance score of model \(M\) on dataset \(D_j^l\), which is derived by randomly shuffling the values of feature \(j\) in \(D\) for the \(l\)-th permutation, and \(L\) is the total number of permutations.
\end{theorem}

\subsection{Principal Component Analysis (PCA)}
Principal Component Analysis (PCA) is a dimensionality reduction technique that transforms a large set of correlated variables into a smaller set of uncorrelated variables known as principal components. The primary goal of PCA is to capture as much variance as possible with the fewest number of principal components. This process involves the following steps:

1. \textbf{Standardization}: The data is standardized to have a mean of zero and a standard deviation of one.
2. \textbf{Covariance Matrix Computation}: The covariance matrix of the standardized data is computed.
3. \textbf{Eigen Decomposition}: The eigenvalues and eigenvectors of the covariance matrix are calculated. The eigenvalues represent the variance captured by each principal component, while the eigenvectors represent the directions of the principal components.
4. \textbf{Principal Components Selection}: The principal components are selected based on their eigenvalues, typically retaining enough components to explain a desired percentage of the total variance (e.g., 90\%).

Mathematically, if \(\mathbf{X}\) is the standardized data matrix, then the covariance matrix \(\mathbf{C}\) is given by:
\[
\mathbf{C} = \frac{1}{n-1} \mathbf{X}^T \mathbf{X},
\]
The eigenvalues \(\lambda_i\) and eigenvectors \(\mathbf{v}_i\) are obtained by solving:
\[
\mathbf{C} \mathbf{v}_i = \lambda_i \mathbf{v}_i.
\]
The principal components are then given by:
\[
\mathbf{Z} = \mathbf{X} \mathbf{V},
\]
where \(\mathbf{V}\) is the matrix of selected eigenvectors.

PCA transforms correlated variables into uncorrelated principal components, effectively reducing multicollinearity in the data. This transformation is particularly useful in regression analysis, where multicollinearity can lead to unstable estimates of regression coefficients.

\subsection{Principal Component Regression (PCR)}
Principal Component Regression (PCR) combines PCA and multiple linear regression. The steps involved in PCR are:

1. \textbf{PCA on Predictor Variables}: Perform PCA on the predictor variables to obtain the principal components.
2. \textbf{Selection of Principal Components}: Select a subset of principal components that explain a sufficient amount of variance.
3. \textbf{Regression Analysis}: Use the principal components selected as predictors in a multiple linear regression model to predict the response variable.

The mathematical formulation of PCR is as follows:

1. Let \(\mathbf{X}\) be the matrix of predictor variables and \(\mathbf{Y}\) be the response variable.
2. Perform PCA on \(\mathbf{X}\) to obtain the principal components \(\mathbf{Z}\).
3. Select the first \(k\) principal components \(\mathbf{Z}_k\) that explain a significant portion of the variance.
4. Fit a linear regression model:
\[
\mathbf{Y} = \mathbf{Z}_k \mathbf{B} + \mathbf{E},
\]
where \(\mathbf{B}\) is the vector of regression coefficients and \(\mathbf{E}\) is the error term.

PCR effectively addresses multicollinearity by using principal components, which are orthogonal to each other, thereby providing stable estimates of the regression coefficients.

\section{Implementation and Fine Tuning}
In this section, we outline the implementation details of Principal Component Regression (PCR) and Support Vector Machine (SVM) with fine tuning. To implement PCR, we first perform PCA on the predictor variables to reduce their dimensionality. The principal components are then used as predictors in a linear regression model. As for SVM, we use the standard grid search with 5-fold cross-validation, where grid search systematically explores a range of hyperparameters to find the optimal model configuration, and 5-fold cross-validation ensures that the performance is evaluated robustly by averaging results across five distinct training and validation splits, helping to mitigate overfitting and improve generalizability.

\subsection{Models' Fine-Tuning}
Hyperparameter tuning is a critical step in the machine learning pipeline, ensuring that the model is optimized for the given data, thereby improving generalization on unseen datasets. To fine-tune our Support Vector Regressor (SVR) model, we embarked on a systematic exploration of the hyperparameter space using Grid Search coupled with 5-fold cross-validation.

\subsubsection{Hyperparameter Space}
The following hyperparameters were considered:
\begin{itemize}
    \item \textbf{Kernel}: Determines the type of hyperplane used to separate the data. We experimented with \texttt{linear}, \texttt{poly}, and \texttt{rbf}.
    \item \textbf{C (Regularization Parameter)}: This parameter trades off correct classification of training examples against maximization of the decision function’s margin. We tested values in the range [0.1,1,10,100,1000,2000,3000,4000,5000].
    \item \textbf{Gamma}: Defines how far the influence of a single training example reaches. We tried both \texttt{scale} and \texttt{auto}.
    \item \textbf{Degree}: The degree of the polynomial kernel function (\texttt{poly}). Evaluated degrees included 2, 3, and 4, although this parameter is disregarded if the kernel isn’t polynomial.
\end{itemize}

\subsubsection{Results}
The Grid Search, combined with 5-fold cross-validation, revealed the optimal hyperparameters for our dataset as:
\begin{itemize}
    \item Kernel: \texttt{poly}
    \item C: 2000
    \item Gamma: \texttt{auto}
    \item Degree: 2
\end{itemize}

This configuration was determined to offer the most promising performance, balancing the bias-variance trade-off and potentially delivering superior results on unseen data.

\subsection{Principal Component Regression (PCR) Performance}
After performing PCA on the predictor variables and selecting the first three principal components, we used these components as predictors in a linear regression model. The results of the PCR model are as follows:

\begin{itemize}
    \item \textbf{Mean Squared Error}: 0.08226991002545095
    \item \textbf{R-squared}: 0.9431715925806526
    \item \textbf{Regression Coefficients}: [[0.36246693 -0.00369905 0.05123607]]
\end{itemize}

The explained variance ratios for the selected principal components are:
\begin{itemize}
    \item \textbf{PC1}: 0.68652744
    \item \textbf{PC2}: 0.15766246
    \item \textbf{PC3}: 0.08149733
\end{itemize}

The principal component loadings for the first three principal components are:
\begin{table}[H]
\centering
\begin{tabular}{lccc}
\toprule
\textbf{Feature} & \textbf{PC1} & \textbf{PC2} & \textbf{PC3} \\
\midrule
Population & 0.288659 & -0.469154 & 0.090625 \\
Surface Area & 0.230376 & -0.082302 & -0.825116 \\
GDP & 0.321209 & 0.391463 & 0.141022 \\
Total Fossil Fuel Consumption & 0.363920 & 0.075377 & -0.016215 \\
Urban Population & 0.344907 & -0.313251 & 0.084428 \\
Electricity Production & 0.259688 & 0.438979 & -0.340387 \\
Agriculture Gross Production & 0.325330 & -0.231430 & 0.242083 \\
Manufacturing & 0.345919 & 0.227373 & 0.260098 \\
Construction Value & 0.339340 & 0.306241 & 0.161256 \\
Number of Livestock & 0.317710 & -0.352596 & -0.124784 \\
\bottomrule
\end{tabular}
\caption{Principal Component Loadings}
\label{tab:pc-loadings}
\end{table}

Principal component regression (PCR) effectively addresses the issue of multicollinearity by transforming the original predictor variables into orthogonal principal components, thereby providing stable and reliable regression coefficients.

\section{Performance Metrics of Regression Models}

In this section, we present the performance metrics of our regression models, trained and tested with data from numerous countries to explore the dynamics between carbon dioxide emissions and a variety of socioeconomic indicators. We have scrutinized two critical metrics for assessing model performance: R-squared and mean squared error (MSE) for an 80/20 training/testing split, along with cross-validated scores to ensure robustness.

\subsection{Support Vector Regression (SVR) Model}

\begin{table}[H]
\centering
\captionsetup{font=large}
\caption{Performance Metrics for SVR Model (80/20 Training and Testing Ratio)}
\label{tab:svr_performance_metrics}
\begin{tabular}{|l|c|}
\hline
\textbf{Metric/Training-Testing Split} & \textbf{80\%/20\%} \\
\hline
R-squared Score & 0.9895 \\
\hline
Mean Squared Error & 0.0152 \\
\hline
\end{tabular}
\end{table}

Observations from the metrics reveal exceptionally high R-squared values for the 80/20 split, averaging at 0.9895. Such strong R-squared values demonstrate the excellent fit of the model to the data set, indicating that the socioeconomic factors considered possess significant predictive power for carbon dioxide emissions. The uniformity of these high R-squared scores in various splits further attests to the stability and reliability of the model.

MSE scores, indicative of the average squared differences between the observed and predicted values by the model, are comparatively low, reaffirming the precision of the model in forecasting carbon dioxide emissions and highlighting its accuracy.

The SVR model demonstrates formidable predictive strength, as evidenced by high R-squared and low MSE scores, underscoring its utility in capturing the complex interplay between carbon dioxide emissions and socioeconomic factors. The performance metrics suggest areas for further methodological refinement, particularly in exploring different data partitioning strategies to enhance prediction accuracy and mitigate overfitting risks.

To illustrate the prediction ability of the SVR model more clearly, Figure 1 shows the prediction results versus the actual value of the 80/20 training and testing split.

\begin{figure}[H]
\centering
\includegraphics[width=0.7\textwidth]{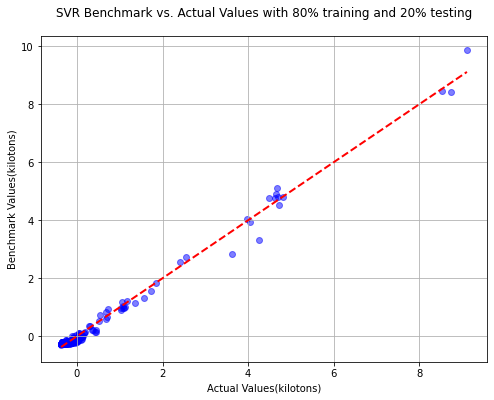}
\caption{SVR Model prediction vs. actual value with 80\% training and 20\% testing}
\end{figure}

From the graph, it is evident that, despite some outliers, most points are closely aligned with the diagonal line, which represents the accuracy of the prediction of 100\%. Thus, the prediction power of the SVR model is reliable.

\subsection{Principal Component Regression (PCR) Model}

To validate the robustness of our PCR model, we employed a k-fold cross-validation. The cross-validated R-squared and MSE scores for the PCR model are presented below:

\begin{table}[H]
\centering
\captionsetup{font=large}
\caption{Cross-Validation Performance Metrics for PCR Model}
\label{tab:pcr_cross_val_metrics}
\begin{tabular}{|l|c|}
\hline
\textbf{Metric} & \textbf{Score} \\
\hline
Cross-Validated R-squared Scores & [0.9428, 0.8421, 0.9174, 0.9215, 0.8827] \\
Mean R-squared & 0.9013 \\
\hline
Cross-Validated MSE Scores & [-0.0828, -0.0858, -0.0981, -0.0927, -0.0711] \\
Mean MSE & 0.0861 \\
\hline
\end{tabular}
\end{table}

The cross-validation results indicate a strong average R-squared score of 0.9013, confirming the reliability of the PCR model across different folds of the dataset. The consistent MSE values further underscore the predictive accuracy and robustness of the model.

To illustrate the prediction ability of the PCR model more clearly, Figure 2 shows the prediction results versus the actual value of the 80/20 training and testing split.

\begin{figure}[H]
\centering
\includegraphics[width=0.7\textwidth]{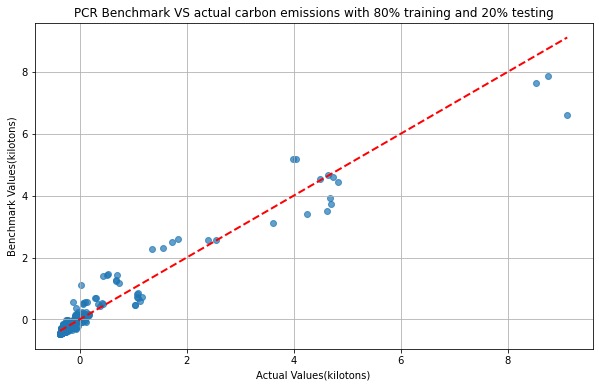}
\caption{PCR Model prediction vs. actual value with 80\% training and 20\% testing}
\end{figure}

From the graph, it is evident that, despite some outliers, most points are closely aligned with the diagonal line, which represents the accuracy of the prediction of 100\%. Thus, the prediction power of the PCR model is reliable.

Next, we employ the permutation importance technique to assess the importance of different factors in our SVR model. As mentioned previously, permutation importance is a model-agnostic feature importance technique that evaluates the impact of individual features on the model’s predictions. It does so by comparing the performance of the model in the original dataset with its performance in terms of the R-squared score after shuffling each feature.

\begin{table}[H]
\centering
\captionsetup{font=large}
\caption{Feature Importance Ranking for 80\%/20\% Training/Testing Split}
\label{tab:model-1-feature-importance-80-20}
\begin{tabular}{ccc}
\toprule
\textbf{Rank} & \textbf{Feature} & \textbf{Importance Value (\%)} \\
\midrule
1 & Total Fossil Fuel Consumption (GWh) & 36.8243 \\
2 & GDP (US \$) & 13.0691 \\
3 & Population & 8.3565 \\
4 & Urban Population & 6.8323 \\
5 & Electricity Production (GWh) & 2.1137 \\
6 & Surface Area (Square KM) & 1.7779 \\
7 & Construction Value (US \$) & 1.3473 \\
8 & Manufacturing (US \$) & 1.2966 \\
9 & Number of Livestock (Heads) & 0.4121 \\
10 & Agriculture Gross Production (million US \$) & 0.2432 \\
\bottomrule
\end{tabular}
\end{table}

From the analysis of feature importances for the 80/20 training/testing split in the SVR model, we observe the dominant influence of 'Total Fossil Fuel Consumption (GWh),', which highlights its pivotal role in predicting carbon dioxide emissions. This consistency underscores the direct impact of a nation’s total fossil fuel usage on its carbon dioxide footprint.

The terms 'population' and 'GDP (US\$)' also consistently rank highly, reinforcing the significant roles these factors play in the estimation of carbon dioxide emissions. The steadfast positions of these features illustrate the undeniable correlation between population size, economic output, and carbon dioxide emissions. These elements serve as fundamental drivers in the models, emphasizing the interplay between demographic scale, economic activity, and environmental impact.

Although the 'urban population' consistently emerges as a crucial factor, its ranking varies slightly, indicating its significant but fluctuating impact on the predictions of carbon dioxide emissions. This fluctuation suggests that while urbanization is a key determinant of carbon dioxide emissions, its relative influence can be modulated by other socioeconomic factors.

'Electricity production (GWh)' and 'Surface area (KM square)' exhibit variable importance, reflecting the nuanced relationship these factors have with carbon dioxide emissions. Electricity production, in particular, showcases how energy generation methods and efficiency levels can significantly influence a country’s carbon dioxide footprint.

Interestingly, “Construction Value (US \$)” and “Manufacturing (US \$)” show a noteworthy presence, pointing to the considerable effect of the industrial sector on carbon dioxide emissions. These factors highlight the environmental cost of industrial and construction activities, underscoring the need for sustainable practices in these areas.

Less prominently ranked features such as “Number of Livestock (Heads)” and “Agriculture Gross Production (million US \$)” still contribute valuable information, suggesting the role of the agricultural sector in carbon dioxide emissions. Although these factors are lower, they underscore the broader spectrum of contributors to a nation’s carbon dioxide emissions, from agriculture to industrial production.

The observed variations in feature importance shed light on the complex interdependencies among socioeconomic, demographic, and environmental factors in the carbon dioxide emission dynamics. These insights call for a deeper exploration into how these variables interact to shape global carbon dioxide emission profiles, providing valuable guidance for targeted policy and intervention strategies to mitigate environmental impact.

\subsection{Comparing Countries' Performance}

\begin{figure}[H] 
    \centering
    \includegraphics[width=\textwidth]{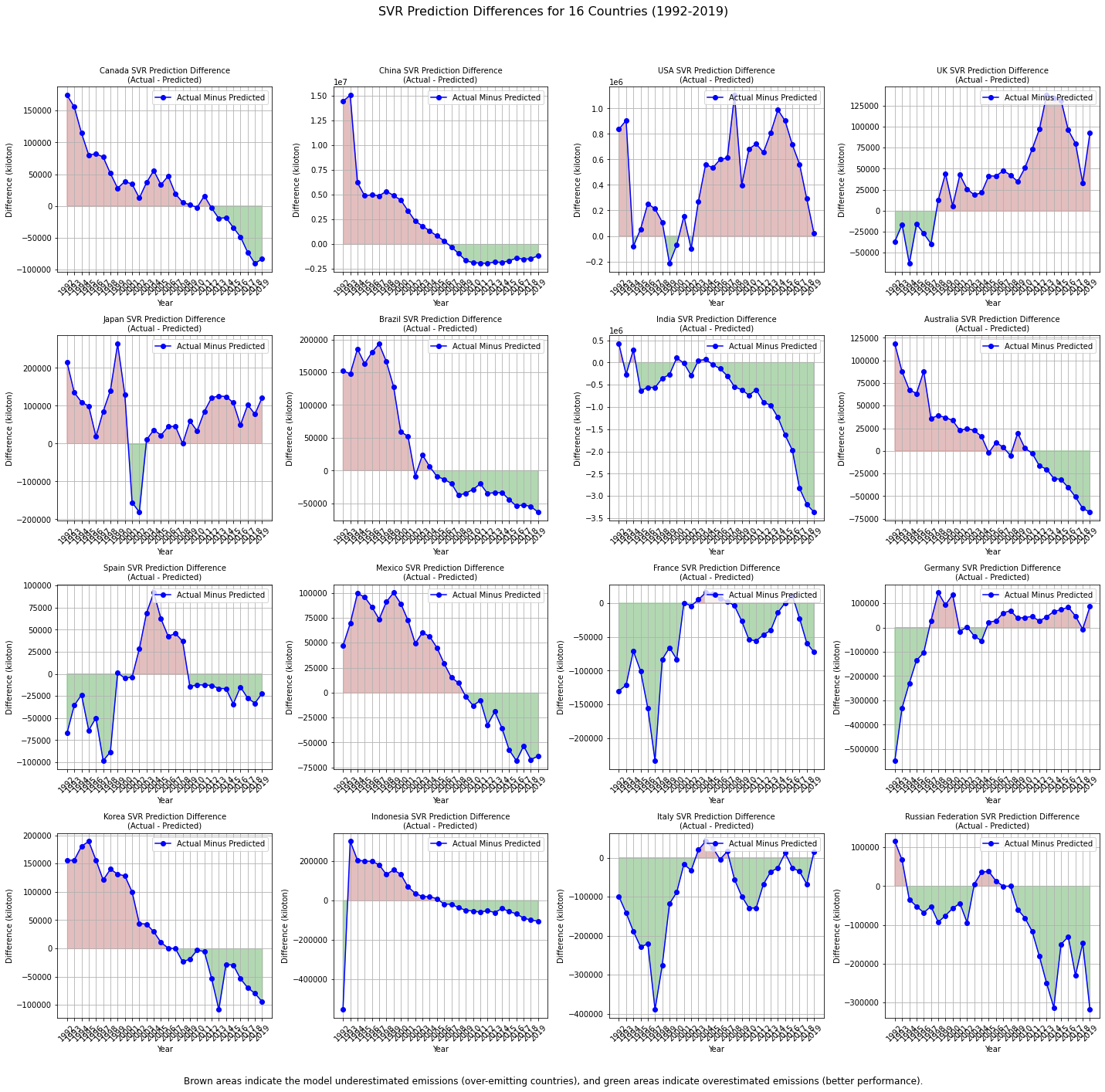}
    \caption{Difference between actual carbon dioxide emissions and SVR model predictions. Brown areas indicate overestimation (predicted emissions higher than actual), while green areas indicate underestimation (predicted emissions lower than actual).}
\label{diffSVR}
\end{figure}

\begin{figure}[H] 
    \centering
    \includegraphics[width=\textwidth]{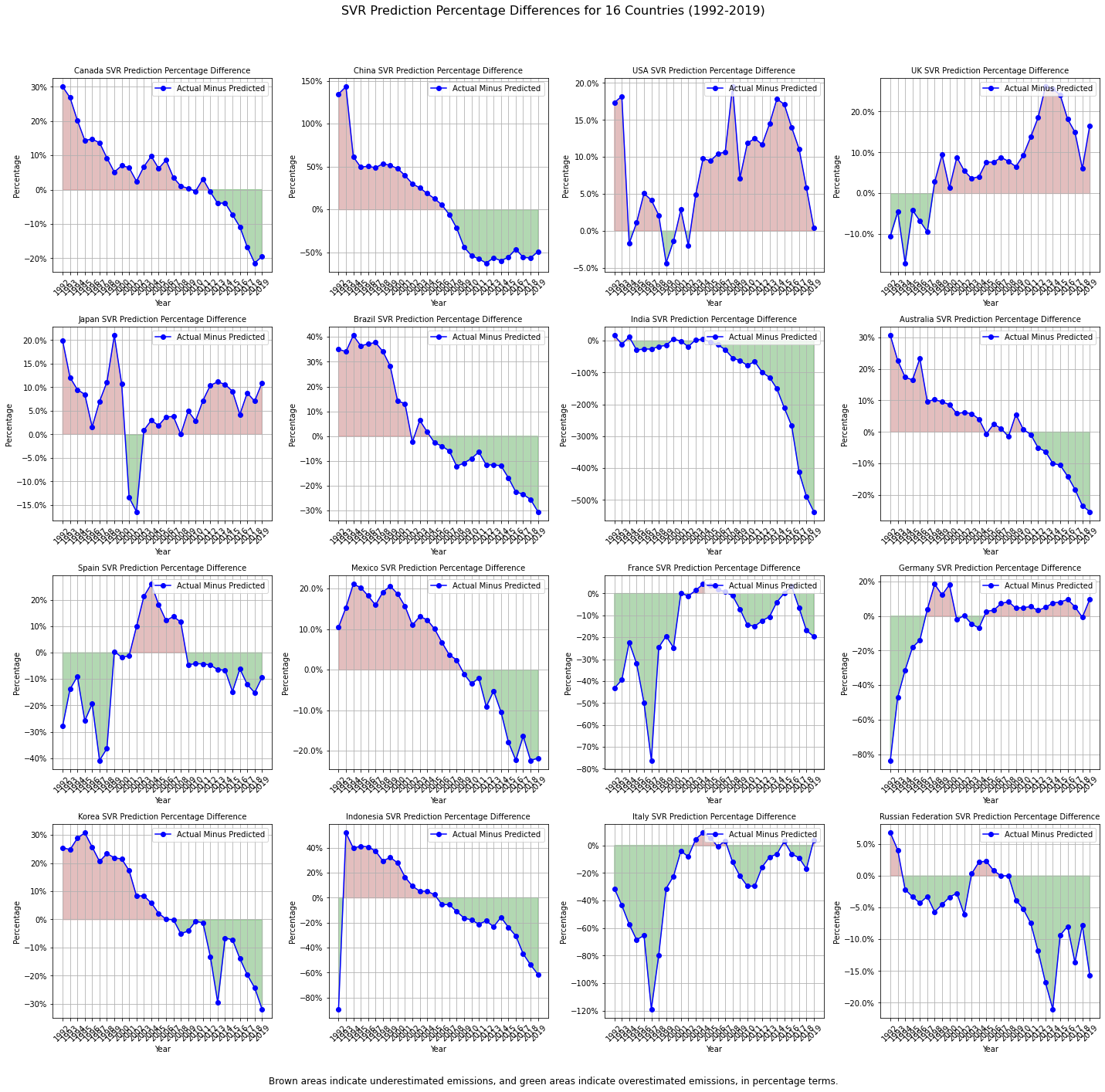}
    \caption{Percentage difference between actual carbon dioxide emissions and SVR model predictions. Brown areas represent overestimation, and green areas represent underestimation. This visualization highlights the relative magnitude of discrepancies across countries.}
\label{percSVR}
\end{figure}

\begin{figure}[H] 
    \centering
    \includegraphics[width=\textwidth]{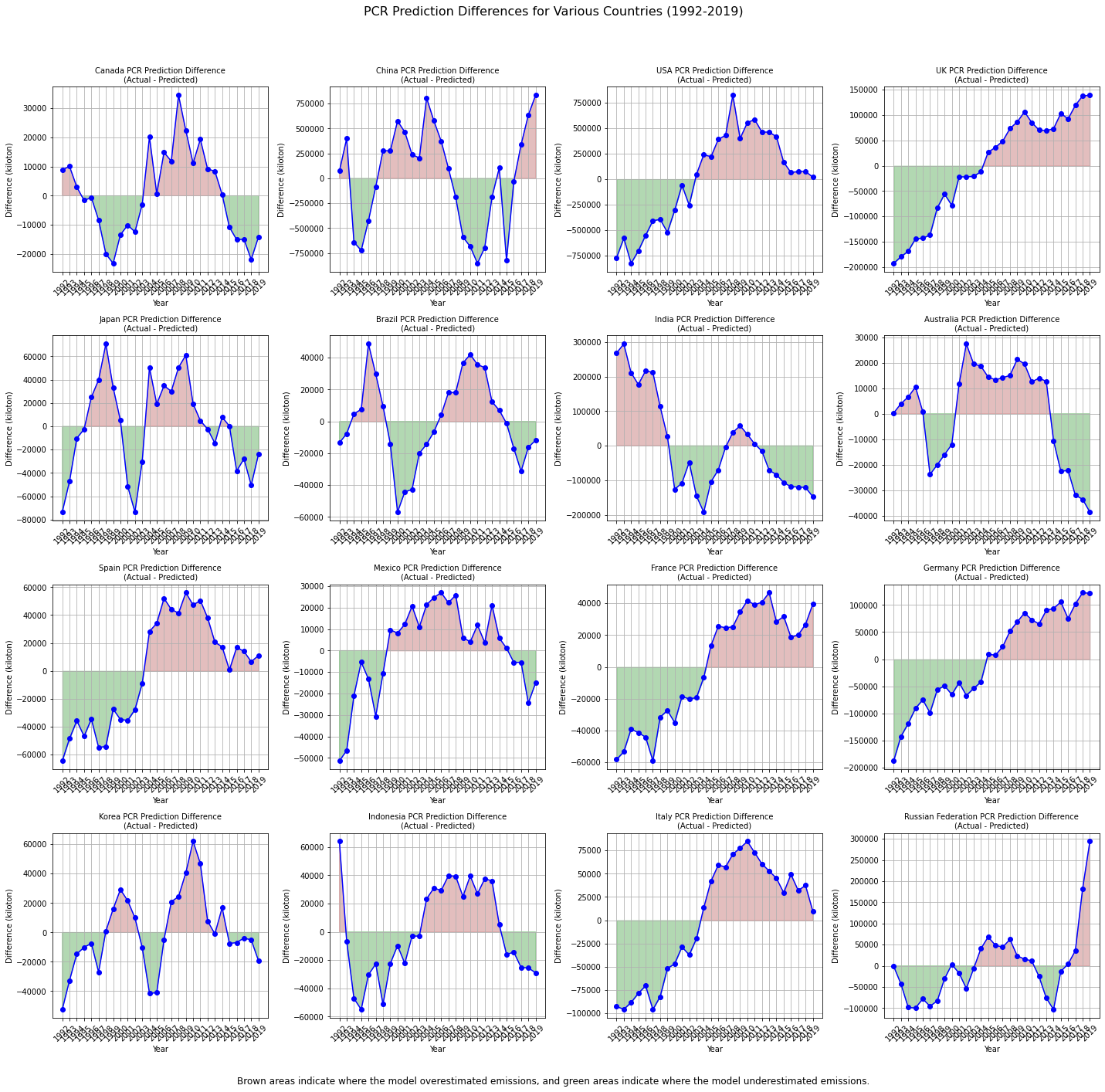}
    \caption{Difference between actual carbon dioxide emissions and PCR model predictions. Brown areas indicate overestimation (predicted emissions higher than actual), and green areas indicate underestimation (predicted emissions lower than actual).}
\label{diffPCR}
\end{figure}

\begin{figure}[H] 
    \centering
    \includegraphics[width=\textwidth]{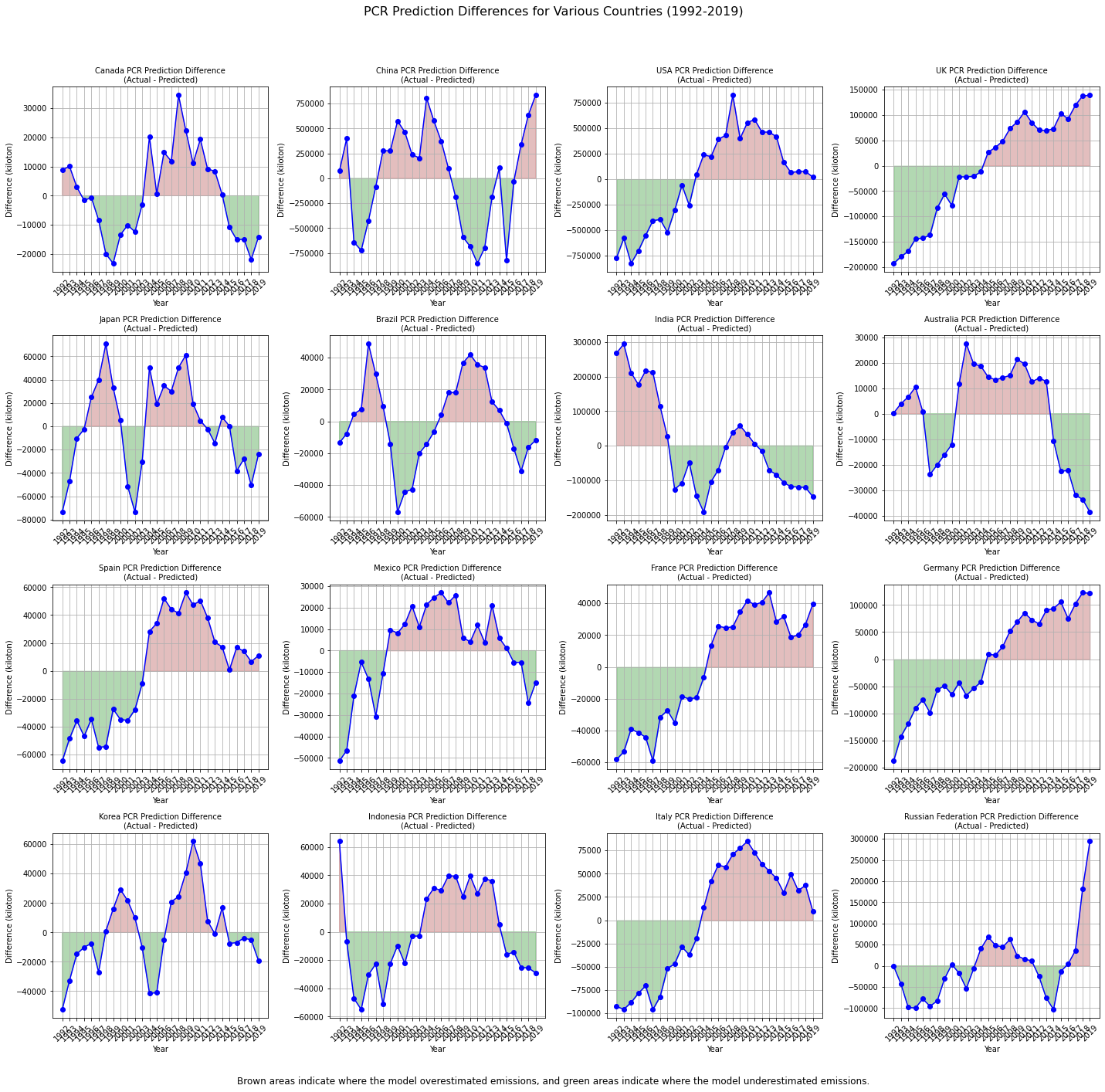}
    \caption{Percentage difference between actual carbon dioxide emissions and PCR model predictions. Brown areas indicate overestimation, and green areas indicate underestimation. The figure illustrates the relative magnitude of errors in prediction.}
\label{PercPCR}
\end{figure}

In our quest to create a predictive model with wide applicability, we trained the model using 80\% of a comprehensive dataset from 1992 to 2019, covering 62 countries. This approach significantly enhanced the model’s ability to predict global carbon dioxide emissions. Focusing on 16 major economies, we analyzed two sets of graphs: the raw differences between predicted and actual emissions, and these differences as percentages relative to actual emissions.

Training with a global dataset enables us to benchmark individual countries’ carbon emission performances against global expectations. This comparison provides insights into whether a country’s efforts align with global standards, offering a valuable framework for evaluating national progress in emission reduction
\subsection{Principal Component Regression (PCR) and Support Vector Regression (SVR) Analysis}

This section evaluates the country-specific trends and predictive accuracy of the PCR and SVR models based on differences between predicted and actual carbon dioxide emissions. Figures \ref{diffPCR} and \ref{PercPCR} (PCR) and Figures \ref{diffSVR} and \ref{percSVR} (SVR) illustrate these discrepancies for 16 major economies, offering complementary insights into global emission dynamics.

The PCR model provides valuable insights into the environmental trajectories of individual nations, capturing both underestimations and overestimations that reflect diverse industrial and policy changes:
\begin{itemize}
    \item \textbf{UK}: The model’s predictions transition from overestimation to underestimation in recent years, likely reflecting evolving factors such as changes in energy sources or industrial activities.
    \item \textbf{Canada}: Early overestimations followed by sharp fluctuations, including downward and upward spikes, indicate significant variability in emission factors. These shifts suggest the impact of effective policy interventions and periods of intensified industrial activity.
    \item \textbf{USA}: Persistent underestimations, with notable spikes in 2009 and 2016, may reflect challenges in capturing the complexities of the country’s emissions profile, including its energy consumption and industrial structure.
    \item \textbf{China}: The model reveals mixed trends, with large early overestimations followed by a steady decline into underestimations. This trajectory aligns with China’s rapid industrialization and recent emission reduction efforts.
\end{itemize}

The SVR model, leveraging its flexibility in capturing non-linear relationships, also identifies significant discrepancies across countries. While it shares some findings with PCR, its ability to model more complex patterns offers a nuanced perspective:
\begin{itemize}
    \item \textbf{USA}: Consistent underestimations suggest gaps in representing the intricate interplay of factors driving emissions.
    \item \textbf{China and Canada}: Early underestimations give way to declining differences, indicating that environmental policies in these nations are progressively mitigating emissions.
    \item \textbf{Australia}: A downward trend from overestimation to underestimation reflects changes in energy policies and industrial behavior.
    \item \textbf{Germany}: Increasing underestimations suggest recent emissions growth, possibly due to shifts in industrial structure or energy sources, which the model does not fully capture.
\end{itemize}

These country-specific trends underscore the complexities of forecasting emissions, emphasizing the importance of combining PCR and SVR to capture diverse dynamics effectively.

\subsection{Insights and Conclusion}

Both PCR and SVR analyses reveal consistent patterns in global carbon dioxide emissions, offering complementary strengths that enhance the overall understanding of emission dynamics. The PCR model excels at addressing multicollinearity and providing stable regression coefficients, while SVR’s non-linear flexibility allows it to adapt to more intricate relationships among factors. Together, these models deliver a robust framework for predicting emissions and identifying key drivers.

Across both models, Total Fossil Fuel Consumption, GDP, and population emerge as the dominant predictors of emissions, confirming the pivotal role of economic and energy consumption factors. Lesser predictors, such as livestock numbers and agricultural output, have relatively minor impacts, reinforcing the dominance of industrial and urban contributors to global emissions. These findings provide a clear basis for policy recommendations aimed at reducing emissions by transitioning to cleaner energy sources and focusing on sustainable economic activities.

The graphical analyses for major economies highlight unique national trajectories. While the USA consistently shows underestimations, reflecting the complexity of its emission profile, Canada and China exhibit fluctuations that align with effective policy interventions. In the UK, the shift from overestimation to underestimation points to evolving industrial and energy patterns, while Australia and Germany display trends that underscore the necessity of refining models to account for recent changes in emission dynamics.

This study demonstrates the value of combining PCR and SVR to benchmark national carbon performance against global standards. By leveraging global datasets and machine learning techniques, these models provide actionable insights for policymakers, enabling targeted strategies to address emission drivers. Future research can build on this work by incorporating additional variables, refining predictive algorithms, and extending the analysis to sector-specific emissions, further enhancing its utility in global sustainability efforts.

\section{Discussion}

This research utilized advanced machine learning techniques, including Support Vector Regression (SVR) and Principal Component Regression (PCR), to develop a robust predictive framework for analyzing carbon dioxide emissions. The models were applied to a dataset spanning 62 countries and covering a wide range of socioeconomic and environmental variables, offering both predictive accuracy and actionable insights.

The SVR model demonstrated exceptional performance, achieving an R-squared value of 0.9895 and a Mean Squared Error (MSE) of 0.0152. These results validate the model’s ability to accurately capture the relationship between emissions and the selected predictors. Its consistent performance across different data splits highlights the stability and reliability of the approach. In parallel, the PCR model effectively addressed the issue of multicollinearity by transforming correlated variables into orthogonal principal components, yielding stable regression coefficients. The model’s mean R-squared of 0.9013 during cross-validation further underscores its utility as a complementary tool to SVR.

Country-specific analysis revealed diverse patterns, reflecting variations in national policies, economic structures, and environmental conditions:
\begin{itemize}
    \item \textbf{United States}: The models consistently underestimated emissions, pointing to the complexity of capturing its diverse emission sources and regional variability.
    \item \textbf{China}: A shift from early underestimations to later overestimations highlights the challenges posed by its rapid industrialization and evolving emission-reduction efforts.
    \item \textbf{Canada}: Fluctuations, with early overestimations followed by periods of underestimation, suggest the dynamic impact of its policies on emissions.
    \item \textbf{United Kingdom}: The models captured a clear transition from overestimation to underestimation, likely reflecting gradual improvements in emissions management.
\end{itemize}

The Permutation Importance analysis identified "Total Fossil Fuel Consumption (GWh)," GDP, and population size as the most influential predictors of carbon dioxide emissions. These results validate the choice of features and provide a roadmap for policymakers to target the most impactful areas for intervention. The ability to benchmark national performances against global trends further enhances the utility of this research, providing a foundation for evaluating and refining emissions policies worldwide.

By integrating SVR, PCR, and feature importance techniques, this study provides a comprehensive understanding of carbon dioxide emissions dynamics. The dual emphasis on predictive accuracy and interpretability ensures that the models offer both practical insights and a scalable framework for future applications.

\section{Conclusion}

This study investigated the global dynamics of carbon dioxide emissions using data from 62 countries, leveraging advanced machine learning techniques to understand the complex relationships between emissions and socioeconomic factors. By employing SVR and PCR, we developed models that not only achieved high predictive accuracy but also provided interpretable insights into the key drivers of emissions.

The SVR model’s R-squared value of 0.9895 and MSE of 0.0152 underscore its effectiveness as a predictive tool, while PCR addressed multicollinearity and produced reliable regression coefficients with a mean R-squared of 0.9013. Together, these models complement each other, offering a robust analytical framework.

Country-specific trends revealed the variability in emission trajectories. The United States’ consistent underestimation points to challenges in modeling its diverse emission sources. In contrast, China’s shift from underestimation to overestimation reflects its rapid industrial growth and subsequent emission reductions. Canada’s fluctuating patterns highlight the dynamic effects of its environmental policies, while the United Kingdom’s transition from overestimation to underestimation signals progressive improvements in managing emissions.

The Permutation Importance analysis highlighted the significance of fossil fuel consumption, GDP, and population size as primary drivers of emissions, reinforcing the importance of these factors in targeted policymaking. By benchmarking national performances against global standards, this research provides actionable insights to guide efforts in reducing emissions and combating climate change.

In conclusion, this study demonstrates the power of combining machine learning techniques with comprehensive datasets to achieve accurate predictions and meaningful interpretations of carbon dioxide emissions. The findings contribute to the broader goal of global sustainability by equipping policymakers with the tools and knowledge needed to address one of the most pressing challenges of our time.

\newpage

\bibliographystyle{unsrt}

\bibliography{mybib.bib}

\end{document}